\documentclass[final,3p]{elsarticle}

\usepackage{amssymb}
\usepackage{xcolor}
\usepackage{adjustbox}
\usepackage{amsmath}

\usepackage{lineno}

\usepackage[utf8]{inputenc}
\usepackage[english]{babel}
\usepackage{fancyhdr}
\usepackage{lastpage}
\usepackage{setspace}
\journal{ }
\begin{document}

\begin{frontmatter}



\title{Reducing Air Pollution through Machine Learning}

\author[inst1,inst2]{Dimitris Bertsimas}
\affiliation[inst1]{organization={Operations Research Center, Massachusetts Institute of Technology},
            city={Cambridge},
            state={MA},
            country={USA}}

\affiliation[inst2]{organization={Sloan School of Management, Massachusetts Institute of Technology},
            city={Cambridge},
            state={MA},
            country={USA}}

\author[inst1,inst2]{Léonard Boussioux}

\author[inst1,inst2]{Cynthia Zeng}

\begin{abstract}
\textbf{Problem Definition:}
This paper presents a data-driven approach to mitigate the effects of air pollution from industrial plants on nearby cities by linking operational decisions with weather conditions. \\
\textbf{Academic Relevance:}
Our method combines predictive and prescriptive machine learning models to forecast short-term wind speed and direction and recommend operational decisions to reduce or pause the industrial plant's production. We exhibit several trade-offs between reducing environmental impact and maintaining production activities. \\
\textbf{Method and Results:} The predictive component of our framework employs various machine learning models, such as gradient-boosted tree-based models and ensemble methods, for time series forecasting. The prescriptive component utilizes interpretable optimal policy trees to propose multiple trade-offs, such as reducing dangerous emissions by 33-47\% and unnecessary costs by 40-63\%. Our deployed models significantly reduced forecasting errors, with a range of 38-52\% for less than 12-hour lead time and 14-46\% for 12 to 48-hour lead time compared to official weather forecasts. We have successfully implemented the predictive component at the OCP Safi site, which is Morocco's largest chemical industrial plant, and are currently in the process of deploying the prescriptive component.\\ 
\textbf{Managerial Insights:}
Our framework provides a pathway for sustainable industrial development by forgoing the trade-off between pollution and industrial activity by linking operational decisions with data-driven weather conditions. This represents a significant step in optimizing factory operations and improving sustainability efforts. As such, it has the potential to modernize how factories approach planning and resource allocation under environmental compliance. Our predictive component has significantly improved production efficiency, allowing for better resource allocation and reduced downtime. This not only led to cost savings for the company but also helped to reduce the environmental impact of production by minimizing air pollution.

\end{abstract}


\end{frontmatter}

\section{Introduction}

Sustainable industrial development is an important issue shared by many countries. The trade-offs between economic activities, environmental pollution, and public health must be managed attentively. Studies show that many environmental toxins have been released into the atmosphere due to urbanization and industrialization over the last 200 years \cite{manisalidis2020environmental, brunekreef2002air, power2018monitoring}. In particular, emissions from chemical power plants can pose significant health risks to those living in the surrounding area \citep{tong_near-source_2015,tong_microenvironmental_2017}.
Therefore, there is a pressing need to develop technologies and infrastructures to achieve economic objectives and environmental preservation simultaneously. 

As data availability and computing methods continue to advance, there has been growing interest in applying machine learning techniques to air pollution management. Previous research has primarily focused on predicting the health consequences of pollution exposure, as demonstrated in a comprehensive review by \cite{bellinger_systematic_2017}.
Additionally, various studies have attempted to forecast air pollution, air quality, and airborne particle concentrations using data such as satellite imagery, weather data, and air quality monitoring data \citep{madan_air_2020, doreswamy_forecasting_2020, castelli_machine_2020, sanjeev_implementation_nodate, kumar_air_2022}.
Despite these efforts, there remains a lack of literature connecting air pollution prediction to decision-making and mitigation actions.
Earlier works on technology-aided tools to reduce pollution include \cite{zhu03}, which discusses a mathematical formulation and algorithm for controlling air pollution using weather forecasts and numerical models to minimize control-related costs, and \cite{panagopoulos_decision_2012}, which proposes a decision support tool to find optimal Best Management Practice locations for minimizing diffuse surface water pollution.

This paper tackles the critical issue of urban air pollution by proposing a novel plant operation scheduling methodology that leverages machine learning and optimization. Our predictive and prescriptive framework links operational decisions to weather forecasts to effectively minimize the impact of air pollution in industrial settings. To the best of our knowledge, our work is the first attempt to reduce industrial air pollution through machine learning. The framework is implemented on the Safi production site of the OCP group in Morocco. 
In summary, our contributions are three-fold: 
\begin{itemize}
\item A data-driven pollution framework incorporating two components: (i) a machine learning-enhanced weather forecasting system that utilizes onsite sensors and official forecasts (ii) an optimization-based operational decision recommendation system optimizing the trade-off between potential pollution risk and operational loss. The predictive component of our framework has been deployed and guides production planning in real-time at the OCP Safi site since July 2022. The prescriptive component is under implementation.
\item Since implementation, our machine learning-enhanced forecasts significantly improved accuracy: we reduced the next 12-hour wind forecasting errors by 38-52\% and the next 12 to 48-hour errors by 14-46\%. In addition, our optimization-based operational decision framework can lead to 40-63\% operational savings while reducing potential polluting cases by 33-47\%.  
\item Our work offers a case study of achieving industrial activities while controlling air pollution's impact on surrounding urban cities. We hope to inspire future work applying machine learning and data science for sustainable industrial development. 
\end{itemize}

\section{Methodology}

This section first provides an overview of the procedure at the Safi factories before this study to manage the emission of dangerous airborne pollutants. Then, we detail how we set up our predictive and prescriptive methodology to improve the pollution reduction pipeline.

\subsection{The Previous Operational Procedure at Safi}

The OCP group is the world's largest phosphate producer, controlling 75\% of the world's phosphate reserves and accounting for more than 30\% of global production. The OCP Safi site was established in the 1970s to produce various phosphates for export.
However, fertilizer production is a known contributor to air pollution, releasing harmful airborne substances such as sulfur dioxide (SO$_2$), sulfur trioxide (SO$_3$), hydrogen sulfide (H$_2$S), and hydrogen fluoride (HF), as well as fine and coarse dust, which can pose serious health risks such as respiratory diseases and cancer \citep{report}.
The site is located 10 km southwest of the Safi city center, with more than 300,000 residents. Due to the geographical location, weather conditions play a critical role in air pollution dispersion. Depending on the wind speed and direction, airborne pollution can be carried into Safi, thus posing a threat to public health and bringing high respiratory and ocular discomfort. In 2013, the  site set up a monitoring procedure to reduce the amount of air pollution in the city with responsive production rates --- and consequently airborne emissions --- depending on the meteorological weather forecasts and real-time on-site wind monitoring system. This procedure schedules production rates and personnel based on next-day weather forecasts. It uses real-time wind monitoring systems to adjust in dangerous weather conditions, ensuring the safety of the surrounding community. 

Before this study, the main bottleneck of the procedure was the gap between meteorological forecasts and real-time conditions, thus leading to unnecessary and costly production shutdowns or missed dangerous weather conditions leading to negative health outcomes. The operators in the Safi production site received operational weather forecasts from the national meteorological agency every 12 hours for the next 48 hours. However, these forecasts are frequently inaccurate because they are calculated at a regional level and come with a 5 to 7 hours lag-time due to the long computational costs of dynamical weather forecasts. As a result, planning activities had no access to real-time forecast information and were sensitive to uncertain weather conditions. 

\begin{figure}     \centering
    \includegraphics[width=0.8\linewidth]{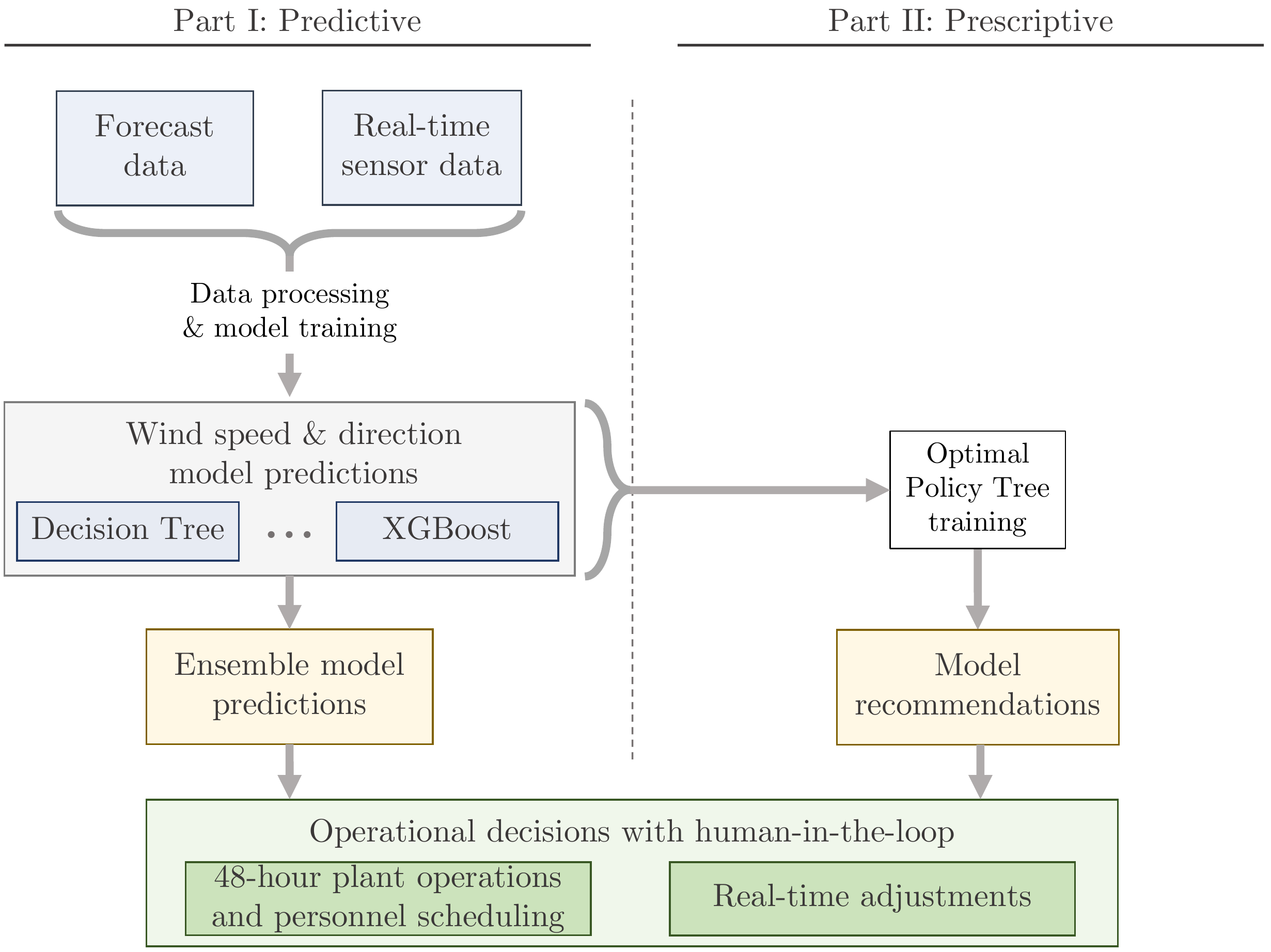}
    \caption{Predictive and prescriptive approach to plant operations scheduling.}
    \label{fig:pipeline}
\end{figure}

This study aims to develop a data-driven framework to reduce the impact of air pollution from industrial plants on nearby cities by responsively adapting production levels based on wind speed and direction.
Our pipeline encompasses two parts: i) machine learning algorithms producing more accurate and frequent wind forecasts by combining official weather data and onsite real-time sensory data to aid short-to-medium term factory and personnel planning; ii) an optimization-based framework to recommend real-time optimal operational decisions taking into account the various forecasts from the machine learning models. Figure \ref{fig:pipeline} illustrates our overarching methodology. 

\subsection{Scenario Definition}

\begin{figure}
    \centering
    \includegraphics[width=0.95\linewidth]{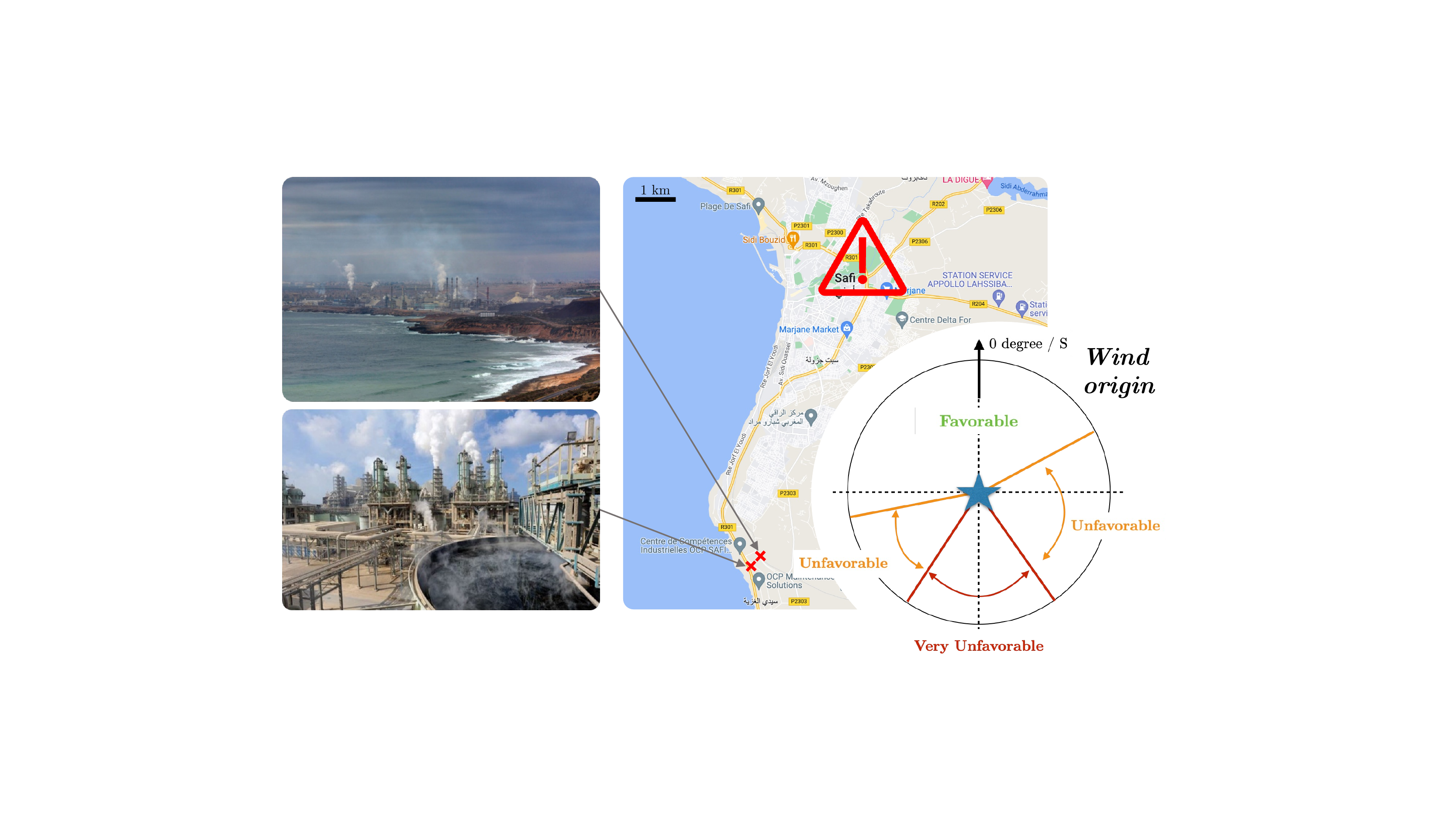}
    \caption{Wind direction and wind speed determine the dissemination of pollutants. Winds coming from the South with low speeds are the most dangerous conditions.}
    \label{fig:plant_location}
\end{figure}

The OCP Safi site developed a warning system to categorize weather conditions into several scenarios to monitor the potential risk of wind carrying pollutants into the nearby city of Safi. 
Scenarios are differentiated as either \textit{favorable} (S1, S2, S2b, S3) or \textit{dangerous} (S3b, S4) based on wind speed and direction, as outlined in Table \ref{tab:scenarios}. This categorization accounts for five wind speed buckets and three wind direction buckets, providing detailed guidance on production rates for each scenario. A dangerous scenario is characterized by low wind speed combined with an unfavorable wind direction, which results in pollutants being directed toward and lingering in the city (see illustration in Figure \ref{fig:plant_location}). Based on the real-time and predicted scenarios, operational decisions are made to reduce air pollution according to the action rules outlined in Table \ref{tab:scenariostype}.

\begin{table}[h] 
\begin{adjustbox}{width=1\textwidth}
\begin{tabular}{|c|c|c|c|}
\hline
\textbf{Wind} & \textbf{Favorable Wind Direction} & \textbf{Very Unfavorable Wind Direction} & \textbf{Unfavorable Wind Direction}\\
\textbf{Speed} & NW, N-NW, N, N-NE, NE, E-NE, E & S-SW, S, S-SE & E-SE, SE, SW, W-SW, W, W-NW \\
\cline { 2 - 4 } (m.s$^{-1}$) & $0^{\circ} - 101.25^{\circ} \text{ \& } 303.75^{\circ} - 0^{\circ}$ & $146.25^{\circ}-$ $213.75^{\circ}$ & $101.25^{\circ}-146.25^{\circ}\text{ \& }213.75^{\circ}-303.75^{\circ}$ \\
\hline $\mathrm{V}<0.5$ & $\mathrm{~S} 3$ & $\mathrm{~S} 4$ & $\mathrm{~S} 4$ \\
\hline $0.5\leq \mathrm{V}<1$ & $\mathrm{~S} 2$ & $\mathrm{~S} 3 \mathrm{~b}$ & $\mathrm{~S} 2 \mathrm{~b}$ \\
\hline $1<\mathrm{V}\leq 2$ & $\mathrm{~S} 1$ & $\mathrm{~S} 3 \mathrm{~b}$ & $\mathrm{~S} 2 \mathrm{~b}$ \\
\hline $2<\mathrm{V}\leq 4$ & $\mathrm{~S} 1$ & $\mathrm{~S} 3 \mathrm{~b}$ & $\mathrm{~S} 2$ \\
\hline $4<\mathrm{V}$ & $\mathrm{~S} 1$ & $\mathrm{~S} 1$ & $\mathrm{~S} 1$ \\
\hline

\end{tabular}
\end{adjustbox}
\caption{Scenario definitions based on wind speed and direction, accounting for five wind speed buckets and three wind direction buckets. Scenarios are differentiated as either favorable (S1, S2, S2b, S3) or dangerous (S3b, S4).}\label{tab:scenarios}
\end{table}

\begin{table}[h]
\centering
\begin{adjustbox}{width=1\textwidth}
\begin{tabular}{|c|c|c|c|}
\hline
\textbf{Scenario} & \textbf{Underlying} &\textbf{Scenario Characteristics} & \textbf{Public Health Consequences} \\ 
\textbf{Type}&\textbf{Scenarios} & &\\
\hline
Favorable & S1, S2, S2b, S3 & High wind speed and/or favorable wind direction & Limited \\
Dangerous & S3b, S4 & Low wind speed and unfavorable wind direction
& Pollutants directed toward and lingering in city\\ \hline
\end{tabular}
\end{adjustbox}
\caption{Categorization of scenarios as favorable and dangerous based on wind speed and direction.}\label{tab:scenariostype}
\end{table}

The effectiveness of the plant's operational response to dynamic weather conditions is contingent on the accuracy of real-time forecasts. If a dangerous scenario is projected in the next three hours, the plant operators proactively adjust production levels accordingly. Once at reduced capacity, production remains at this level until a favorable scenario is predicted and real-time weather conditions become favorable. This highlights the critical importance of precise weather forecasting, as any inaccuracies can result in costly consequences for production and personnel scheduling.

Before our work, plant managers used official forecasts as mere guidance and often relied on experience due to the low accuracy in near-term predictions. This led to frequent inconsistencies and last-minute adjustments under unforeseen weather conditions, underscoring the need for real-time, high-quality weather forecasts in informed decision-making.

\subsection{Predictive Methodology}

Our predictive framework focuses on developing machine learning models to produce accurate hourly wind forecasts by integrating official weather data and onsite real-time sensory data to aid short-to-medium factory and personnel planning. Since July 2022, our predictive framework has been implemented at the OCP Safi site, resulting in reduced production downtime, improved resource allocation, and cost savings. This successful implementation serves as a model for other factories seeking to improve their sustainability efforts and reduce their environmental impact.

\paragraph{Data Processing}

We combined two datasets to make predictions: the official weather forecast data and real-time weather measurement data collected with on-site sensors. Data used in this study range from July 2015 to March 2022. 

Official forecasts are received twice daily around 6:00 am (GMT) and 6:00 pm (GMT) from the Moroccan National Meteorological Department. 
These forecasts are produced by traditional dynamical models with initial conditions and often take 5-7 hours of computational time. 
They provide hourly values for the next 48 hours for wind speed, wind direction, humidity, solar irradiance, and temperature at the Safi site. We call this model the \emph{baseline model} in the rest of the paper.
The on-site sensors measure the same five weather features (wind speed, wind direction, humidity, solar irradiance, and temperature) at one-minute intervals. 

We first imputed the missing values caused by electronic or server malfunctions with linear interpolation. We then averaged the measurement data over one-hour intervals. We used the arithmetic average for the humidity, solar irradiance, and temperature, and the vector average technique \citep{grange} for wind speed and direction (e.g., the vector average of a southerly and a northerly wind of 5 m.s$^{-1}$ gives a mean wind speed of 0 m.s$^{-1}$ because there is no resultant wind speed).  
We encoded the wind direction using the cosine and sine transformations to avoid singularities at endpoints due to the cyclical nature of the feature.

\paragraph{Training data creation}

We transformed the time-series data into a standard tabular form to train traditional machine-learning models. To make wind predictions at time $t$ for the hour $t+h$, we concatenated the present and past 48 hours of weather measurement features at each time step into a vector. Then, we appended the following features: the latest operational forecast available at time $t$ for wind speed, wind direction, pluviometry, and solar irradiance; the cosine and sine of the day and the hour corresponding to time $t$. Table \ref{tab:features} summarizes the 304 features and associated processing techniques. 

\begin{table}[h]
\centering
\begin{adjustbox}{width=1\textwidth}
    
\begin{tabular}{|l|l|l|l|} 
\hline
\textbf{Feature Description} & \textbf{Processing Technique} & \textbf{Initial} & \textbf{Number of} \\
&&\textbf{Feature Range} & \textbf{features}\\
\hline
Wind speed & Vector average & 0.00 - 14.20\ m.s$^{-1}$ & 49\\
Wind direction & Vector average, cos/sin encoding & [0, 360]$^{\circ}$ & $49\times2$ \\
Solar irradiance & Arithmetic average & 0.0 - 978.4 W.m$^{-2}$ & 49 \\
Temperature & Arithmetic average &  4.8 - 46.7$^{\circ}$C & 49\\
Pluviometry & Arithmetic average & 0.0 - 17.2 mm & 49 \\
Day of the year & Cos/sin encoding & 1 - 365 & 2\\
Hour of the day & Cos/sin encoding & 0 - 23 & 2\\ 
Official forecast for wind speed &  & 0.0 - 16.5 m.s$^{-1}$& 1 \\
Official forecast for wind direction & Cos/sin encoding & [0, 360]$^{\circ}$ & 2 \\
Official forecast for pluviometry &  & 0.0 - 20.8 mm & 1 \\
Official forecast for solar irradiance &  & 0 - 1074 W.m$^{-2}$ & 1 \\
Official forecast for temperature &  & 3.2 - 43.1$^{\circ}$C & 1 \\
\hline

\end{tabular}
\end{adjustbox}

\caption{Table recording all the features and processing techniques. The number of features obtained accounts for concatenating the past 48-hour values.}

\label{tab:features}
\end{table}

\paragraph{Model Training}

For the prediction task, we trained five different types of machine learning models to predict wind speed and direction, including Elastic Net, Decision Trees, Random Forest, LightGBM, and XGBoost. To handle the cyclical property for wind direction, we predicted the cosine and sine of the angle instead of the raw angle degree. Predictions are then converted back into scenario predictions using Table \ref{tab:scenarios}. We trained one model for each lead time between 1 and 48 hours ahead, i.e., $48 \times 3$ regression models for wind speed, cosine, and sine of wind direction. For each model, we performed hyperparameter tuning using the validation set as explained later in Section \ref{sec:trainingprotocol}. 

In addition, we trained ensemble models to predict wind speed and direction for every lead time using predictions from these previous individual machine-learning models. Ensemble modeling is a well-established technique to leverage the strengths and limitations of multiple models and benefit from their diversity. The principle is to combine the predictions of the forecasting models available to obtain a more accurate, stable, and robust predictor. In our case, we used the stacking \citep{wolpert} concept and tried several ensemblers, including decision trees, regularized linear regression, and gradient-boosted trees. Elastic Net regression performed the best, and we considered it as our final ensemble model technique.

\subsubsection{Qualitative Feedback from Real-World Implementation}

Our collaboration with OCP's software development team has resulted in the seamless integration of our weather forecasts into the company's internal system. As of July 2022, the site manager and plant operators have been utilizing the forecasts produced by our framework through a simple user interface. They check the hourly forecasts before scheduling production shutdowns, leading to a significant reduction in production downtime.

Qualitative feedback from production managers has indicated that our forecasts are substantially more accurate than official weather forecasts and provide valuable real-time updates that are particularly advantageous during winter when wind conditions are more unpredictable. This has improved factory planning and resource allocation, allowing for more efficient production, better personnel scheduling, and cost savings for the company.

The successful implementation of our framework at OCP Safi is a testament to our approach's effectiveness in optimizing factory operations. We believe that utilizing our framework has the potential to advance how factories approach planning and resource allocation, ultimately leading to improved sustainability efforts and environmental impact reduction.

\subsection{Prescriptive Methodology}

The management team at OCP Safi recognizes the importance of taking immediate action in response to dangerous weather conditions. As a result, our focus is on utilizing short-term (next 3 hours) weather predictions to inform plant operations. To achieve this, we employed Optimal Policy Trees (OPT) \citep{amram} to determine the most optimal decision in real time given the forecasts made by the different machine learning models. Despite the imbalanced nature of the data, with dangerous scenarios accounting for only 1.5-2\% of the total observations, our prescriptive models aim to balance the trade-off between financial savings and effective pollution management. This approach is crucial to mitigate the potential for false negatives in predictive models.
 
In our context, the prescriptive approach deals with observational data of the form $\{(\mathbf{x}_i, y_i, z_i)\}$. Each observation $i$ consists of features $\mathbf{x}_i \in \mathbb{R}^{18}$ (the ensemble members' predictions), an applied prescription $z_i\in \{0,1\}$ (reduce plant production or not), and an observed outcome $y_i\in \mathbb{R}$ (real-world costs associated with the decision). Our prescriptive task is determining the optimal policy that, given the features $\mathbf{x}$, prescribes the treatment $z$ that results in the best outcome $y$. The prescription involves choosing between one of two available decisions, either to reduce production or not.

Table \ref{tab:outcomes} outlines the reward matrix utilized to train the Optimal Policy Tree and quantifies the costs associated with false positives and false negatives. First, no cost is incurred if the forecasted scenario and actual conditions are favorable. When the plant operates at reduced levels as a conservative measure after forecasting a dangerous scenario, the factory incurs a loss of earnings of \$2,000 per hour due to decreased production and the expenses of injecting odor control chemicals to minimize unpleasant odors in the surrounding area. On the other hand, the failure to forecast a dangerous scenario leads to the plant operating at a normal level and polluting the nearby city when the weather conditions turn dangerous. Afterward, the plant operators need to shut down production urgently and inject odor control chemicals. We propose evaluating various public health costs ranging from \$2,000 to \$18,000. This parameter yields differing trade-offs between pollution and costs and can be determined based on the decision-makers' level of conservatism and risk aversion. 

\begin{table}
\centering 
\begin{adjustbox}{width=1\textwidth}

\begin{tabular}{|c|c|c|c|c|}
\hline \textbf{Forecasted} & \textbf{Actual} & \textbf{Cost}  & \textbf{Decision} & \textbf{Public Health}\\
\textbf{Scenario}& \textbf{Scenario} & (USD)& \textbf{Outcome} &\textbf{Impact}\\
\hline Favorable & Favorable & 0 & Full level production & Low\\
Dangerous & Favorable & 2000 & Reduced level production + anti-odor injection & Low \\
Dangerous & Dangerous & 2000 & Reduced level production + anti-odor injection & Low\\

Favorable & Dangerous & 4000 - 20000 & Full production before urgent shutdown + anti-odor injection & High \\

\hline
\end{tabular}
\end{adjustbox}
\caption{Reward matrix for training the Optimal Policy Trees based upon forecasted and actual weather conditions.}\label{tab:outcomes}
\end{table}

\subsection{Training Protocol}\label{sec:trainingprotocol}

The data covers August 2015 to March 2022, totaling 43,952 hourly samples. The data set was divided into training (60\%), validation (20\%), and testing (20\%) sets. The validation set was used to tune the hyperparameters of the machine-learning models. The ensemble models and optimal policy tree parameters were 5-fold cross-validated on the predictions made on the validation set. All models were evaluated on the unseen test set corresponding to the real-world deployment phase.

\begin{table}[h]
\centering
\begin{tabular}{|l|l|l|}
\hline
\textbf{Model} & \textbf{Hyperparameters} & \textbf{Values} \\
\hline
Elastic Net & regularization $\alpha$ coefficient & 0.2, 0.4, 0.6, 0.8, 1 \\
& $\ell_1$ ratio & 0.5, 0.7 \\
\hline
Decision Trees & maximum tree depth & 5, 6, 7, 8, 9, 10 \\
& minimum samples per split & 3, 5, 7 \\
& minimum samples per leaf &  4, 6\\
\hline
Random Forest 
& bootstrap & True, False \\ 
& number of estimators & 100, 150 \\
& maximum tree depth & 5, 6 \\
& min samples split & 4, 6 \\
\hline
LightGBM 
& number of leaves & 31, 60 \\
& maximum tree depth & 4, 6\\
& learning rate &  0.1, 0.3 \\
& lambda $\ell_1$ & 0, 1 \\
\hline
XGBoost & 
number of estimators & 100, 150  \\
& maximum tree depth & 4, 6\\
& learning rate &  0.1, 0.3 \\
\hline
Elastic Net Ensemble & regularization $\alpha$ coefficient & 0.2, 0.4, 0.6, 0.8, 1 \\
& $\ell_1$ ratio & 0, 0.25, 0.5, 0.75, 1.0 \\
\hline
\end{tabular}
\caption{Hyperparameters searched for our models.}
\label{tab:hyperparameters}
\end{table}

\paragraph{Software Tools}

We used Python 3.8 \citep{python} and the scikit-learn package \citep{scikit} to implement all machine learning models. We used the Python package InterpretableAI \citep{iai} to train Optimal Policy Trees. 

\section{Results}

The predictive methodology results reported in this section are the real-world results achieved at the Safi factories after deployment in December 2020. The prescriptive methodology results are back-tested during the same period as the deployment is underway.

\subsection{Predictive Methodology Results} 

Tables \ref{tab:resultspeed} and \ref{tab:resultangle} report the results of the wind speed and wind direction forecasting tasks for all the regression models we deployed at the Safi site: the baseline model, Elastic Net, Decision Tree, Random Forest, Light GBM, XGBoost, and the Elastic Net ensemble model. For each wind prediction task, we report the mean absolute error (MAE) and the expected shortfall at 85\%, corresponding to the average error on the worst 15\% samples. 

In general, all machine learning models improve upon the baseline model, with XGBoost and Light GBM being the best two methods. In addition, the ensemble model further improves the MAE and expected shortfall, especially in near-term horizons. Looking at speed prediction, for less than 12-hour lead time, the best-performing machine learning approaches can outperform the baseline model by 40\%-50\% in both metrics. For longer-term predictions with more than a 12-hour horizon, the best-performing machine learning approaches can outperform the baseline model by 20-30\%. We observe a similar trend for angle prediction: machine learning approaches can achieve 30-50\% improvement upon the baseline model for less than 12-hour lead time predictions and 10-20\% improvement for longer lead time predictions. 

In addition, we observe that the ensemble model outperforms the best single machine learning model consistently across tasks and error measures. The advantage of an ensemble model is especially strong for less than 12-hour lead time predictions. The ensemble model can achieve 0-8\% MAE reduction depending on the specific lead time (except for a slightly worse performance on the longer-term expected shortfall for speed).

\begin{table}[!ht]
    \centering
    \begin{adjustbox}{width = \textwidth}

    \begin{tabular}{|c|c|c|c|c|c|c|c|c|}
    \hline
        \textbf{Lead Time} & \textbf{Metric} & \textbf{Baseline} & \textbf{Elastic Net} & \textbf{Decision Tree} & \textbf{Random Forest} & \textbf{LightGBM} & \textbf{XGBoost} & \textbf{Ensemble} \\ 
        \hline
        1 & ~ & 0.96 & 0.54 & 0.56 & 0.53 & 0.49 & 0.49 & \textbf{0.48} \\ 
        2 & ~ & 1.05 & 0.71 & 0.76 & 0.71 & 0.64 & 0.65 & \textbf{0.63} \\ 
        3 & ~ & 1.18 & 0.80 & 0.85 & 0.8 & 0.72 & 0.72 & \textbf{0.70} \\ 
        6 & MAE & 1.61 & 0.91 & 0.97 & 0.91 & 0.85 & 0.85 & \textbf{0.84} \\ 
        12 & (m.s$^{-1})$ & 1.94 & 1.0 & 1.05 & 0.99 & 0.96 & 0.96 & \textbf{0.94} \\
        24 & ~ & 1.37 & 1.07 & 1.11 & 1.08 & 1.06 & 1.06 & \textbf{1.04} \\ 
        36 & ~ & 2.13 & 1.17 & 1.20 & 1.17 & 1.16 & 1.16 & \textbf{1.15} \\ 
        48 & ~ & 1.57 & \textbf{1.17} & 1.22 & 1.18 & \textbf{1.17} & \textbf{1.17} & \textbf{1.17} \\ \hline
        1 & ~ & 2.37 & 1.40 & 1.48 & 1.36 & 1.27 & 1.28 & \textbf{1.26}  \\ 
        2 & ~ & 2.60 & 1.80 & 1.95 & 1.77 & 1.63 & 1.65 & \textbf{1.61}  \\ 
        3 & Expected & 2.94 & 2.01 & 2.16 & 1.99 & 1.81 & 1.82 & \textbf{1.78}  \\ 
        6 & Shortfall  & 3.82 & 2.27 & 2.45 & 2.25 & 2.15 & \textbf{2.14} & \textbf{2.14}  \\
        12 & 85\% & 4.55 & 2.48 & 2.64 & 2.44 & \textbf{2.38} & 2.40 & 2.39  \\ 
        24 &  & 3.51 & 2.61 & 2.77 & 2.60 & \textbf{2.58} & \textbf{2.58} & 2.59  \\ 
        36 & ~ & 4.93 & 2.79 & 2.89 & 2.76 & \textbf{2.75} & \textbf{2.75} & 2.77  \\ 
        48 & ~ & 4.03 & 2.80 & 2.95 & 2.79 & \textbf{2.78} & \textbf{2.78} & 2.81 \\ \hline
    \end{tabular}
    \end{adjustbox}
    \caption{Beta test results on the test set for wind speed prediction for all models. We record the MAE and expected shortfall at 85\% level for different lead times ranging from 1 hour to 48 hours.}
    \label{tab:resultspeed}
\end{table}

\begin{table}[!ht]
    \centering
     \begin{adjustbox}{width = \textwidth}
        
    \begin{tabular}{|c|c|c|c|c|c|c|c|c|}
    
     \hline
        \textbf{Lead Time} & \textbf{Metric} & \textbf{Baseline} & \textbf{Elastic Net} & \textbf{Decision Tree} & \textbf{Random Forest} & \textbf{LightGBM} & \textbf{XGBoost} & \textbf{Ensemble} \\  \hline
        1 & ~ & 25 & 26 & 14 & 13 & \textbf{12} & 13 & \textbf{12} \\  
        2 & ~ & 26 & 31 & 20 & 18 & 17 & 17 & \textbf{16} \\  
        3 & ~ & 29 & 35 & 23 & 21 & 19 & 19 & \textbf{18} \\  
        6 & MAE & 41 & 40 & 29 & 28 & \textbf{24} & \textbf{24} & \textbf{24} \\  
        12 & (m.s$^{-1}$) & 58 & 43 & 35 & 33 & 31 & 31 & \textbf{30} \\  
        24 & ~ & 42 & 45 & 40 & 38 & 37 & 38 & \textbf{36} \\  
        36 & ~ & 70 & 49 & 45 & 42 & 42 & 42 & \textbf{41} \\  
        48 & ~ & 52 & 48 & 47 & 44 & 44 & 44 & \textbf{42} \\ \hline
        1 & ~ & 73 & 82 & 59 & 53 & \textbf{51} & \textbf{51} & \textbf{51} \\  
        2 & ~ & 79 & 103 & 77 & 72 & \textbf{68} & \textbf{68} & \textbf{68} \\  
        3 & Expected & 89 & 117 & 90 & 83 & 76 & 75 & \textbf{74} \\  
        6 & Shortfall & 115 & 132 & 108 & 107 & \textbf{94} & 95 & 95 \\  
        12 & 85\% & 136 & 138 & 127 & 125 & \textbf{117} & \textbf{117} & \textbf{117} \\  
        24 & ~ & \textbf{127} & 143 & 138 & 135 & 134 & 135 & 132 \\  
        36 & ~ & 155 & 148 & 145 & \textbf{140} & \textbf{140} & \textbf{140} & \textbf{140} \\  
        48 & ~ & 145 & 148 & 146 & \textbf{143} & \textbf{143} & \textbf{143} & \textbf{143} \\ \hline 
    \end{tabular}
    \end{adjustbox}
    \caption{Beta test results on the test set for wind direction prediction for all models. We record the MAE and expected shortfall at 85\% level for different lead times ranging from 1 hour to 48 hours.}
    \label{tab:resultangle}
\end{table}

\subsection{Prescriptive Methodology Results}  

Table \ref{tab:prescriptive} compares the performance of several models for recommending binary hourly actions (anticipating dangerous conditions or maintaining production levels). It includes the baseline model, the previous Elastic Net ensemble model, and a series of Optimal Policy Trees (OPT) with different health costs associated with false negatives (-4000, -6000, -10000, -15000, and -20000). For each model, the table reports the number of false positives and false negatives, the percentage of unnecessary costs that can be saved, and the dangerous pollution reduction compared to the baseline. The unnecessary costs correspond to the amount of money lost by reducing the production and injecting anti-odor chemicals while weather conditions were actually favorable. The dangerous pollution corresponds to the cases when the factory did not reduce production nor injected anti-odor chemicals on time.

The results indicate that as the health cost chosen to train the OPT increases, false negatives decrease, and false positives increase, as expected. Comparing the OPT models with the baseline model shows that the OPT models have overall better performance, as they have lower health costs and fewer false positives. However, it should be noted that there is a trade-off between money savings and pollution reduction. Therefore, the decision-maker must consider their specific goals and priorities when choosing between these objectives.

There are several excellent cost savings / dangerous pollution reduction trade-offs, including (89\%,7\%), (63\%,33\%), (40\%,47\%). In other words, the prescriptive approach is a strict benefit compared to the previous Safi pipeline with the baseline model. The Elastic Net ensemble substantially reduced the unnecessary costs by 82\% but with a 21\% worse pollution outcome than the baseline.

\begin{table}[!ht]
\centering
\begin{adjustbox}{max width=\textwidth}
\begin{tabular}{|l|c|c|c|c|c|}
\hline
\textbf{Model} & \textbf{Health Cost} & \textbf{False Positives} & \textbf{False Negatives} & \textbf{Cost savings} & \textbf{Pollution Reduction} \\
\hline
Baseline &  & 288 & 110 & 0\% & 0\% \\
\hline
Ensemble &  & 51 & 133 & 82\% & -21\% \\
\hline
 & -4000 & 20 & 113 & 93\% & -3\% \\

Optimal & -6000 & 32 & 102 & 89\% & 7\% \\

 Policy & -10000 & 106 & 74 & 63\% & 33\% \\

Tree & -15000 & 174 & 58 & 40\% & 47\% \\

 & -20000 & 282 & 38 & 2\% & 65\% \\
\hline
\end{tabular}
\end{adjustbox}
\caption{Performance of three families of models for recommending actions. We include the baseline model, the Elastic Net ensemble, and a series of optimal policy trees trained with different health costs associated with false negatives.}
\label{tab:prescriptive}
\end{table}

A prescriptive approach, such as the OPT models presented in the table, offers greater flexibility and control in decision-making by allowing for the explicit consideration of trade-offs between different objectives. On the other hand, a predictive approach like the Elastic Net ensemble focuses primarily on providing predictions and does not provide the same level of control and flexibility. 

In addition, the OPTs provide interpretable insights on how the different ensemble members are used to prescribe, as illustrated by Figure \ref{fig:opt} below. In particular, we notice that a simple tree like the one corresponding to choosing a health cost of \$15,000, can reduce pollution emissions during dangerous scenarios by 47\% and save 40\% of the unnecessary costs. Conveniently, it also relies on only three ensemble members: XGBoost and Elastic Net predicting speed, and Random Forest predicting the cosine component of the wind direction. It also suggests that different ensemble members capture different aspects of the data and together make better recommendations. 

\begin{figure}[ht]
\begin{minipage}[b]{0.4\linewidth}
\centering
\includegraphics[width=4.5in]{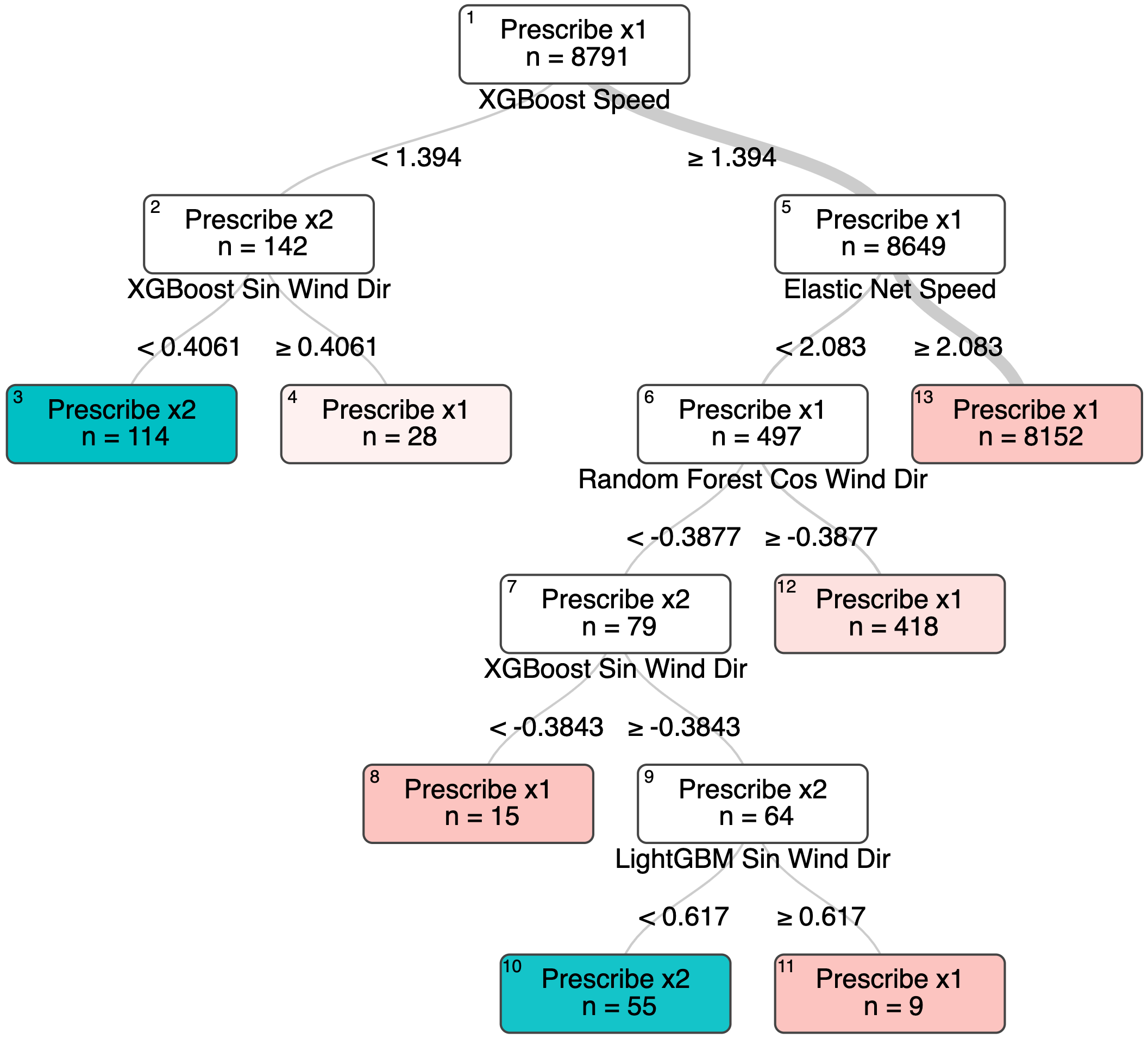}
\end{minipage}
\hspace{3.2cm}
\begin{minipage}[b]{0.5\linewidth}
\centering
\includegraphics[width=2.5in]{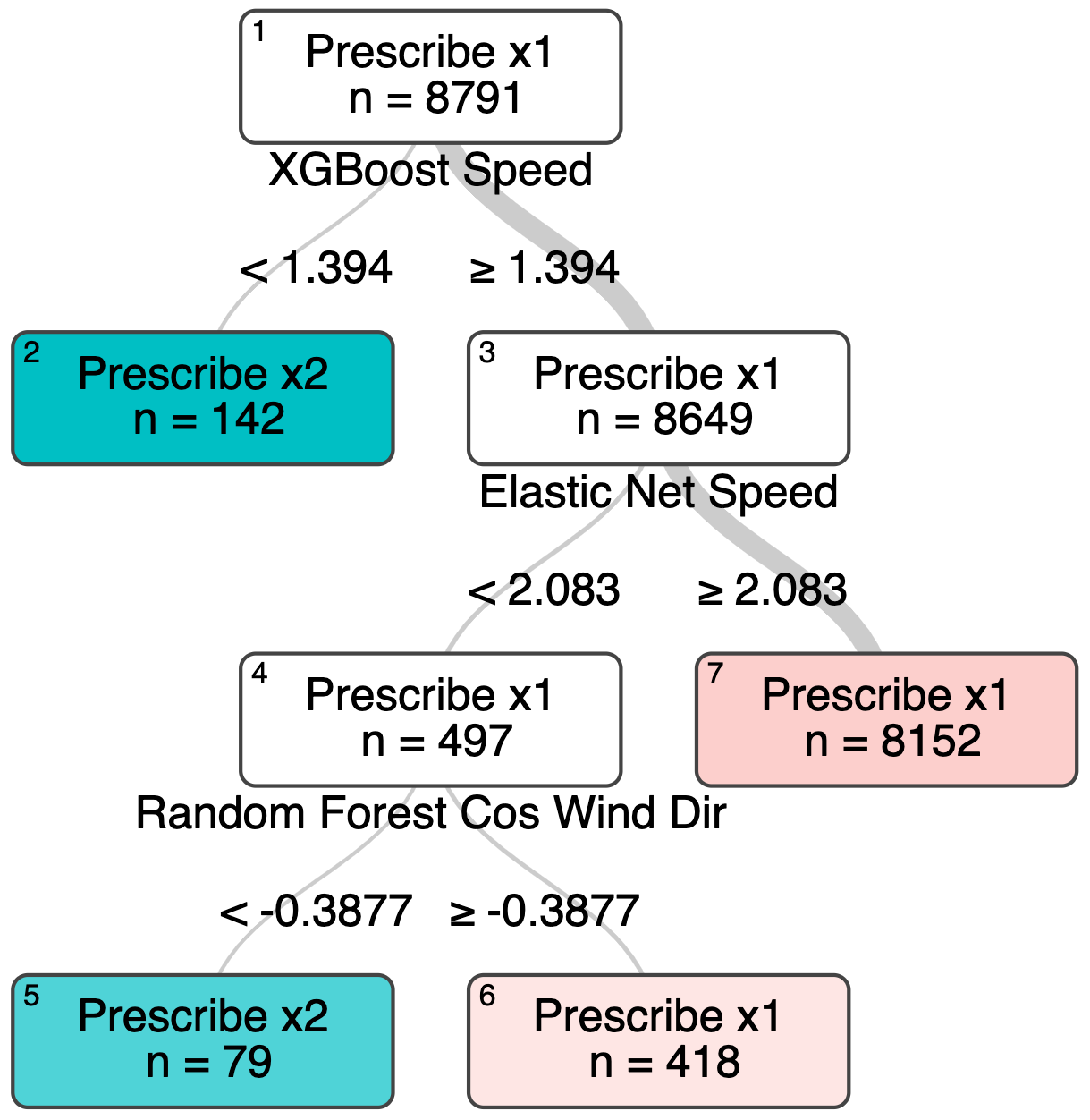}

\end{minipage}
\caption{Optimal policy trees trained using a health cost of \$10,000 (left) and \$15,000 (right). The trees illustrate an interpretable decision-making process to arrive at certain recommended decisions (prescription options).  \texttt{Prescribe x1} corresponds to maintaining production rate while \texttt{Prescribe x2} corresponds to reducing plant operations and injecting odor control chemicals.}\label{fig:opt}

\end{figure}

\section{Conclusion}

In conclusion, our study introduces a novel and data-driven solution to mitigate the harmful effects of air pollution caused by industrial plants in urban areas. We provide a comprehensive solution for managing industrial operations and weather-related risks by combining advanced weather forecasting and decision-making models. Our framework, which incorporates both predictive and prescriptive machine learning models, was successfully implemented at the OCP Safi production site, resulting in improved forecasting accuracy and decision-making efficiency. Given the crucial role of weather in industrial environmental impact, we believe that our approach can be adapted and effectively applied in similar settings.

Our framework has demonstrated its value in managing air pollution in chemical production sites, and the results achieved at the OCP Safi site hold the potential to inspire a more sustainable and responsible chemical production industry globally. The flexibility and adaptability of our approach enable its core components of data enhancement, real-time monitoring, and prescriptive models to be universally applied to different chemical factories. Although each production site presents unique challenges, our data-driven approach can be customized to meet the needs and conditions of each location. Utilizing the latest advancements in weather forecasting and data analysis, we aim to assist factories in effectively managing air pollution and promoting the safety and well-being of the surrounding communities.

\section*{Acknowledgements}
We thank the OCP group for their invaluable contribution to this work. We warmly acknowledge Karim Bendioub, Yassine Azhari, Aimad Aneddame, and Walid Daou, whose support was critical in developing and implementing this work. Karim Bendioub made a major contribution by building the software to integrate the models into the pipelines at the OCP Safi factories and replicating the code behind the firewall.

\bibliographystyle{elsarticle-num} 
\bibliography{cas-refs}

\end{document}